\documentclass[10pt, a4paper]{article}
\usepackage{lrec2022}
\usepackage{multibib}
\newcites{languageresource}{Language Resources}
\usepackage{graphicx}
\usepackage{tabularx}
\usepackage{soul}
\usepackage{algorithm}
\usepackage{algpseudocode}
\usepackage{titlesec}
\titleformat{\section}{\normalfont\large\bfseries\center}{\thesection.}{1em}{}
\titleformat{\subsection}{\normalfont\SmallTitleFont\bfseries\raggedright}{\thesubsection.}{1em}{}
\titleformat{\subsubsection}{\normalfont\normalsize\bfseries\raggedright}{\thesubsubsection.}{1em}{}
\renewcommand\thesection{\arabic{section}}
\renewcommand\thesubsection{\thesection.\arabic{subsection}}
\renewcommand\thesubsubsection{\thesubsection.\arabic{subsubsection}}

\usepackage{epstopdf}
\usepackage[LGR,T1]{fontenc}
\usepackage[utf8]{inputenc}
\newcommand{\textgreek}[1]{\begingroup\fontencoding{LGR}\selectfont#1\endgroup}

\usepackage{hyperref}
\usepackage{xstring}

\usepackage{color}

\title{The Construction and Evaluation of the LEAFTOP Dataset of Automatically Extracted Nouns in 1480 Languages}

\name{Greg Baker, Diego Molla-Aliod} 

\address{Macquarie University \\
         4 Research Park Drive \\
         gregory.baker2@hdr.mq.edu.au, diego.molla-aliod@mq.edu.au \\
         \\}

       \abstract{ The LEAFTOP (language extracted automatically from
         thousands of passages) dataset consists of nouns that appear
         in multiple places in the four gospels of the New
         Testament. We use a naive approach --- probabilistic
         inference --- to identify likely translations in 1480 other
         languages. We evaluate this process and find that it provides
         lexiconaries with accuracy from 42\% (Korafe) to 99\%
         (Runyankole), averaging 72\% correct across evaluated
         languages. The process translates up to 161 distinct lemmas
         from Koine Greek (average 159). We identify
         nouns which appear to be easy and hard to translate, language
         families where this technique works, and future possible
         improvements and extensions. The claims to novelty are: the
         use of a Koine Greek New Testament as the source language;
         using a fully-annotated manually-created grammatically parse
         of the source text; a custom scraper for texts in the target
         languages; a new metric for language similarity; a novel
         strategy for evaluation on low-resource languages.  \\ \newline
         \Keywords{leaftop, bible, corpus, Koine
           Greek, lexicon, low-resource languages} }

\begin{document}

\maketitleabstract

\section{Introduction}

This paper discusses a large new dataset designed to be useful for
tasks involving part-of-speech identification, grammar morphology and
language similarity measures which was created by extracting
vocabulary from Bible translations.
It establishes a baseline of accuracy for target language noun
extraction using sensible naive techniques at scale, for the languages
spoken by the majority of the world's population.




\begin{figure}[t]
\begin{center}
\includegraphics[scale=0.4]{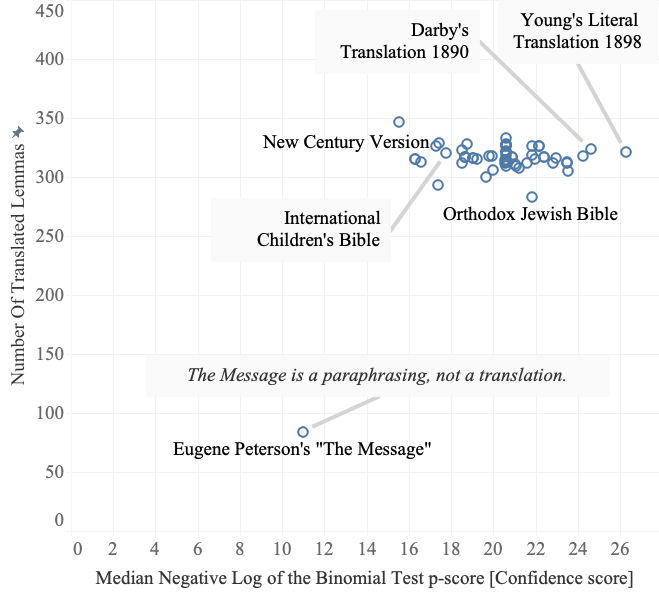}

\caption{Paraphrased vs literal translations of the New Testament into English}
\label{paraphrase}
\end{center}

\medskip
\hrule

\end{figure}

The resulting LEAFTOP dataset is both wide (1480 languages) and deep
(160 nouns). For inflected languages it contains forms for number
(singular vs plural) for all nouns; for some nouns it can also supply
gender and case variations.

\begin{table}
  {\small

  \begin{tabularx}{\columnwidth}{lX}
    {\bf Language Family} & {\bf Languages Evaluated} \\
Niger-Congo     & Fon; Guinea Kpelle; Igbo; Samia; Luganda; Mano; Runyankole; Swahili; Twi; Soga; Yoruba \\
 Afro-Asiatic   & Tunisian Arabic; Modern Standard Arabic; Moroccan Arabic; Chadian Arabic \\
 Dravidian      & Telugu \\
 Artificial     & Esperanto \\
 Arnhem         & Gunwinggu \\
 Austronesian   & Cebuano; Dobu; Hiligaynon; Hiri Motu; Kilivila; Nyindrou; Takia; Tagalog \\
 Trans-New Guinea & Korafe; Melpa \\
 Indo-European   & Bengali; German; French; Hindi; Marathi; Sinhala; Urdu \\
 Nilo-Saharan    & Teso \\
\end{tabularx}
}
  \caption{The subset of the 1480 languages in the LEAFTOP dataset which have
    been evaluated, and the language families they are part of.}
  \label{language-families-evaluated}

  \medskip
  \hrule
\end{table}

\begin{figure}[t]
  \includegraphics[scale=0.4]{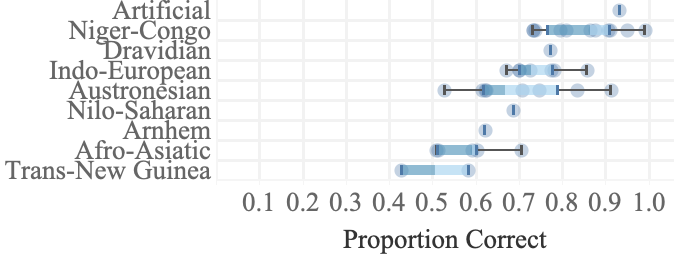}
  \caption{The LEAFTOP extraction approach works surprisingly
    well for languages in the Niger-Congo family and for some Austronesian languages, and less well for Trans-New Guinea
  languages.}
\label{proportion-correct}
\medskip
\hrule
\end{figure}

The Ethnologue
\cite{ethnologue} reports the existence of
7,139 living languages.  There are surprisingly few data
sets which cover a large proportion of the world's population that
also have sufficient depth for machine learning techniques.  There are only
82 nouns listed in the extended Swadesh 207 list; the ASJP database
\cite{asjp} covers only the Swadesh 100 and is therefore even smaller
with only 46 nouns. This is a vocabulary that is too small for many
machine learning techniques. Panlex \cite{panlex} has sufficient depth and
structure for many tasks, but unfortunately 
has surprising gaps for the grammar morphology task we were
pursuing\footnote{For example, as of the time of writing,
  Panlex correctly offers ``mfuasi'' as a translation for the English
  word ``disciple'', but without a direct English-to-Swahili
  translation, it offers ``chama'' (which is incorrect) as a two-step
  translation for ``disciples''.}.

Bible translations exist for the native languages of 80\% of the
world's population \cite{wycliffe_for_lrec}. As of the end of December 2020,
the Bible has been fully translated into 717 languages. A further 1,582
languages have a New Testament translation. As discussed in section
\ref{results}, the four gospels alone are sufficient to generate
singular and plural forms for 160 nouns in most (but not all)
of these 2,299 languages.

\section{Extensions of past work}

Compared to past work in lemma extraction and massively parallel
language corpora, a number of small incremental changes have been made:
\begin{itemize}
\item More successful scraping by writing site-specific parsers.
\item An algorithm (algorithm \ref{alg:writtenstructure}) for identifying whether a language uses word markers and whether
  it has an alphabet.
\item Automated identification of whether the translation attempted to
  be literal or used paraphrasing extensively.
\item Used Koine Greek as the source language and leveraging existing
  manually-created part-of-speech annotations.
\end{itemize}

\newcite{mccarthy-etal-2020-johns} presented their
work collating the Bible in 1600 languages at LREC in 2020, which
re-used the CMU Wilderness Corpus \cite{cmu-wilderness}. They
encountered limitations with this --- verse alignment was challenging
because the verse numbers are in-line in the text. Our
improvement over this approach was to scrape from Bible gateways
where the verse information is encoded in the metadata of
the HTML\footnote{Prior to  the 9th Circuit's decision on
    hiQ Labs, Inc. v. LinkedIn Corp
    in April 2022 scraping of this nature was of dubious legality.
}.

\begin{figure}[t]
\begin{center}
\includegraphics[scale=0.5]{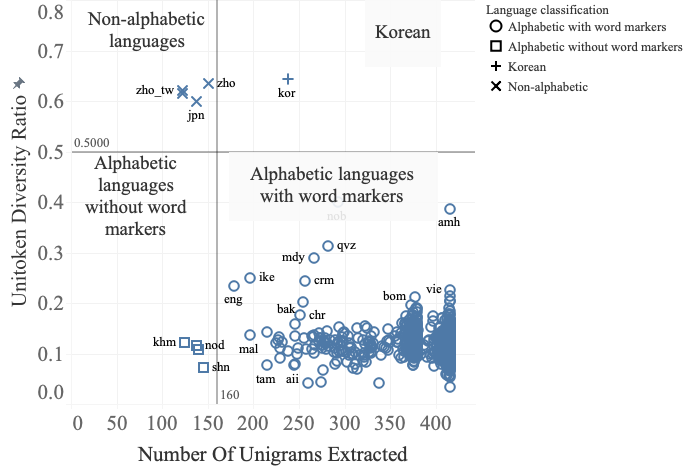}
\end{center}
\caption{Identifying the written structure of the language.
  Korean is included as an alphabetic language with word break markers
by algorithm \ref{alg:writtenstructure} but is --- as would be expected ---
an outlier. All other languages in the dataset are handled correctly.
}

\label{lorth}

\medskip
\hrule

\end{figure}

In addition, where McCarthy et al. could only handle languages that
have word-marker boundaries, the LEAFTOP dataset is able to
distinguish between languages that have word marker boundaries,
languages that don't have word marker boundaries that are nonetheless
alphabetic (e.g. Thai, Khmer), and languages that don't have word
marker boundaries that are non-alphabetic (e.g. Chinese characters,
which LEAFTOP calls ``uni-token'' extractions) using the method in
algorithm \ref{alg:writtenstructure} (discussed in section
\ref{alg-written-structure-writeup}.  For alphabetic languages without
word markers, the LEAFTOP dataset defaults to using quad-tokens, which
is incorrect, and serves as a placeholder for future improvements.

\begin{algorithm}
  \caption{Vocabulary extraction algorithm}\label{alg:vea}
  \begin{algorithmic}

  \State $L \gets$ the set of Greek lemmas that appear
  twice or more in the Gospels maintaining
 the same case, number and gender

  \State $B \gets$ the Bible versions in the target language 

\ForAll{$l \in L$}

\ForAll{$b \in B$}

   \State $C \gets \{$ verses present in translation $b \}$
   
   \State $V \gets C \cap \{$ verses where lemma $l$ appears $\}$

   \State $U \gets C \cap \{$ verses where lemma $l$ does not appear and
   neither does $l$ in any other case, number or gender  $\}$

   \For{$k \gets $unigram, unitoken, quadtoken}

      \State $T \gets \{\}$

      \For{$v \gets $V}

         \State $s \gets$ \Call{TokenizeVerse}{$b,v,k$}

         \State $T \gets T \cup s$

      \EndFor

      \State $m \gets \infty$

      \State $n \gets \{\}$

      \For{$t \gets T$}

      \State $Q \gets$ the verses where $t$ is present

      \State $R \gets$ the verses where $t$ is not present

      \State $X \gets |V \cap Q|$

      \State $Y \gets |V \cap R|$

      \State $Z \gets |U \cap Q|$

      \State $W \gets |U \cap R|$

      \State $p \gets $ \Call{BinomTest}{X,Y,Z,W,greater}

      \If{$p = m$}

      \State $n \gets n \cup \{t\}$

      \ElsIf{$p \le m$}

      \State $n \gets \{t\}$

      \State $m \gets p$

      \EndIf

      \EndFor

      \If{$|n| = 1$}

      \State Trans($b,l,k$) $\gets t$

      \EndIf

      \EndFor

      \EndFor

   \EndFor
\end{algorithmic}
\end{algorithm}

\newcite{christodouloupoulos2015massively} assembled a corpus of
translations into 100 different languages, preferring the oldest
common translation where multiple translations exist. Their hope was
that this would not be too archaic in language and also be the most
literal translation. This led to them choosing the King James
Version\footnote{Which, uniquely among documents from the 17th century
  is not in the public domain. It is protected (still) by royal
  charter, but is allowed to be used for research purposes.}  (or
Authorized Version) for English which sadly fails on both
criteria\footnote{Even at the time of publishing, ``you'' was
  displacing ``thee'' in spoken English for example. The many
  mistranslations of the KJV are well known, see
  \cite{tsoraklidis2001mistranslations} for a small sample.}.  Our
reservations on this choice led us to want to use
statistical methods to identify the translations which are
likely to be literal, using the consistency of the translations of
each source lemma. If two translations into the same language have
substantially different confidence scores\footnote{As discussed in section \ref{confidence-score-definition},
  the confidence score is the ratio of the negative log p-value
  of the best candidate word to the second best candidate word.
} in lemma identification, and also
substantially different numbers of lemmas that can be translated, then
it is clear which version is a paraphrase.  The results of this
for English language translations is shown in Figure \ref{paraphrase}.
These less literal translations are still included in LEAFTOP anyway.

With the exception of translations for the benefit of the Assyrian
Church of the East\footnote{Who hold that the correct source for New
  Testament translations is the Peshitta.}, New Testament translations
are supposed to be translations from Koine Greek. In practice, Bible
translators use a variety of language sources; but when a word has
ambiguous meanings, best practice is to refer to the Greek form.


This may not completely resolve the matter if the Greek itself covers
multiple meanings in the target language. But we would
expect --- in general --- word alignment
from Greek should outperform word alignment from English\footnote{
  Given how translation is done in low-resource languages, word alignment
  for low resource languages will probably work best when the source
  is the dominant language in the local region.}.
Previous attempts to generate vocabulary from New Testament
translations (such as the University of North Texas' submission to
\cite{sigmorphon2019} and \cite{nicolai-yarowsky-2019-learning}) have
used English language sources to recover target language lemmas. And
indeed, these have suffered from low accuracy, with the latter
reference reporting 57.6\% accuracy in their task. This falls
far short of proving the superiority of aligning with Koine Greek in general
for all languages, but
it was sufficient evidence to support the authors' choice for LEAFTOP.

%

Finally, there is a loss of accuracy derived from automated tagging of
parts of speech. An extensive search of the literature has failed to
find any research on how accurate automated POS tagging of English
language Bible translations is beyond what was in  \cite{agic-etal-2015-bit} which
includes the unquantified sentence ``Bible translations typically have
fewer POS-unambiguous words than newswire.'' 
It is unlikely, though, that an automated tagger will be able to
compete with the centuries of analysis that grammarians and
translators have done on the New Testament source documents.

\section{Data sources and Extraction}

The source code is available\footnote{{\small \url{https://github.com/solresol/thousand-language-morphology}}}.  This section
explains the code and the choices made in developing it.
Most of the works that were scraped are held under copyright and
cannot be shared as part of this dataset.

\subsection{Koine Greek Parsing}

\begin{figure}[t]
  \includegraphics[scale=0.25]{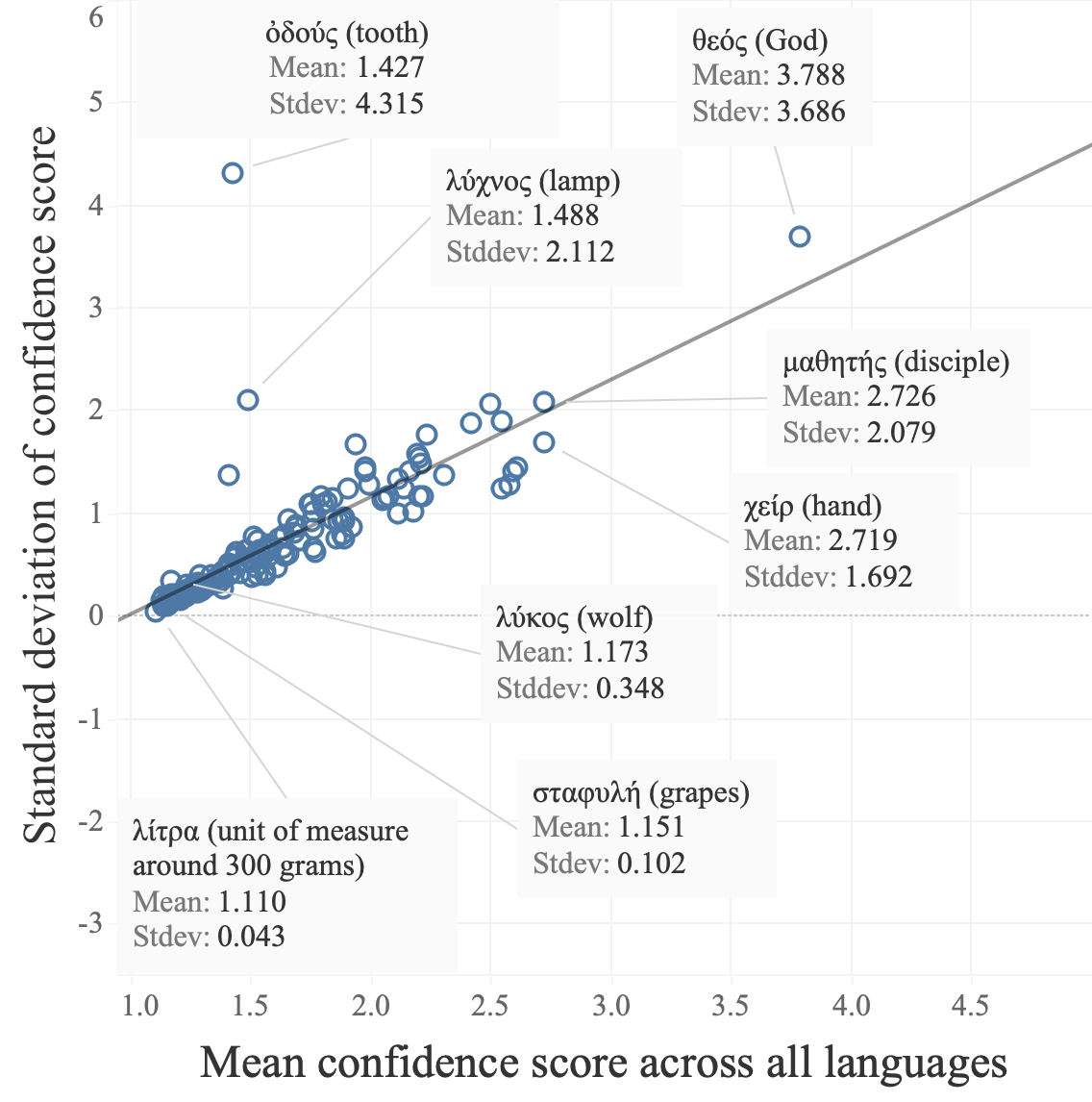}
  \caption{Scatterplot of mean and standard deviation (across all languages) of the confidence
    of the vocabulary extraction for each
    Koine Greek lemma (with English translation)}
  \label{mean-vs-standdev-confidence}

  \medskip
  \hrule
\end{figure}

We chose to extract vocabulary from only the
four Gospels in the hope that they would be using less abstract
vocabulary and --- written for an agrarian society --- have terms that
were mostly universal. This is unfortunately not true, as we
discovered when the translations were evaluated in
Gunwinggu\footnote{Also known as Bininj Gun-Wok and Kunwinjkuan, it
  is a language spoken by the Bininj people of West Arnhem Land in
  northern Australia, who have been mostly nomadic.} where words
like ``governor'' and ``prison'' were untranslatable.

The history of the OpenText.org website is given in
\cite{porter_evans_pitts_2017}; it is a manually-annotated copy of the
Codex Sinaiticus performed by experienced linguists and
theologians. As discussed in \cite{metzger}, the Codex Sinaiticus is the
second-best text we have of the 4 Gospels of the New Testament, but
the grammatical annotations of the Codex Sinaiticus (and the Codex Sinaiticus itself)
are the most accessible to researchers --- they are available on
github \cite{gntannot}. Our review of the annotations found
few mistakes: \textgreek{συκῆ} (fig tree) is marked as neuter in Mark
11:13 instead of feminine; \textgreek{ἀλάβαστρον} (jar for alabaster)
is marked as feminine in Mark 14:3 instead of neuter. On no occasions
was a noun marked as any other part of speech, nor any other part of
speech marked as a noun. This is far from a perfect evaluation, but if
these are the only problems, then the accuracy of these annotations is
above 99.9\%. The OpenText.org annotations also include Louw-Nida
domains \cite{louwnida}, which could be used in future projects for
establishing $p$-adic word embeddings.

The XML sources were imported and all nouns whose annotated lemma did
not begin with a capital letter (979 distinct lemmas%
%
) were identified and grouped based on gender (masculine, feminine,
neuter), number and case (nominative, accusative, dative and
genitive). Koine Greek (by the time of the New Testament) had lost
the dual as a number, and only had singular and plural. Nouns that
appeared only once in a given gender, number and case were dropped
(leaving 567 lemmas%
, in 1185 %
forms).

We made the decision to limit the extraction 
to pairs of nouns that appear in
both singular and plural forms, of which there are only 188.
They appear in 666\footnote{Which is an amusing coincidence.}
different forms. On the one hand, this does allow for interesting
grammar morphology tasks and substantially reduced the computation time
required, but on the other hand, it is an arbitrary constraint
that could be dropped in a future version of the dataset.

In practice, Algorithm \ref{alg:vea} was never able to extract more
than 161 lemmas. On average, it extracted 159.2 terms (standard
devation = 4.87).


\begin{algorithm}[t]
  \caption{Algorithm for identifying the written structure of the language and choosing the
  correct tokenisation method}\label{alg:writtenstructure}
\begin{algorithmic}
  \State $u \gets max(|\{$ Trans($b,l,$unigram) $\forall l \in L\}| \forall b \in B)$
  \State $v \gets max(|\{$ Trans($b,l,$unitoken) $\forall l \in L\}| \forall b \in B)$
  \State $w \gets$ the number of tokens in Trans($b,l,$unitoken) counting duplicates
  \If{$u \ge 160$} 
  \State
  \Return{\tt \scriptsize alphabet,word markers,unigram}
      \ElsIf{$v \ge \frac{w}{2}$}
      \State \Return{\tt \scriptsize non-alphabetic, unitoken}
   \Else
      \State \Return{\tt \scriptsize alphabetic, no word markers, quadtoken}
     \EndIf
   \end{algorithmic}

   $L$ and $B$ are from algorithm \ref{alg:vea}. 160 was found empirically from the
   clustering shown in Figure \ref{lorth}.
\end{algorithm}


\subsection{Scraping}

There are 3370 verses in the Gospels that contain nouns; only 2724 of them
contain one of the 188 lemmas, so a small optimisation is not to
fetch verses that will not be useful. This also helped establish that
the purpose of the scraping is for research and not to create a
complete copy. The scraper was written to use Selenium \cite{selenium_for_lrec} to control a
web browser to fetch the data; allowing for the inefficiency of this,
and long delays to avoid triggering CAPTCHAs, the process of scraping
the 6,008,134 verses from \url{www.bible.com} took several weeks.

The scraper rejected 217 of the 2,356 Bible versions because of some
fundamental problem --- a required book of the Bible being not present being
the most common. Disputed verses (such as the passage from John 7:51 --
8:11, which was merged into the gospels later and therefore not
present in all Bible translations) were captured as empty
strings. This does not appear to reduce the accuracy of the lemma
identification of the 10 nouns that appear in that passage that are
also found in the repeated nouns list, since even the least common
lemma 
\textgreek{δάκτυλος} (finger) appears another 8 times in the Gospels.

\begin{figure}
  \includegraphics[scale=0.25]{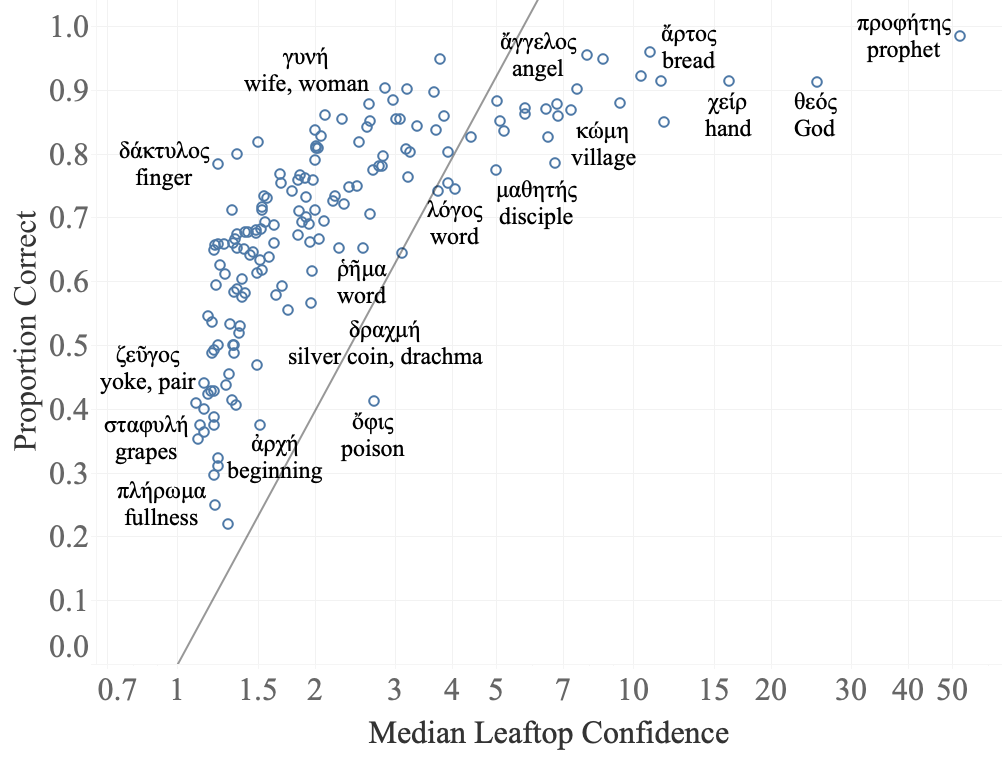}

  Proportion Correct = 0.572*$\ln$(Median Leaftop Confidence).
  P-value < 0.0001, $R^2$ = 0.732
  \caption{Relationship between confidence and accuracy with forced linear regression}
  \label{hardness}
  \medskip
  \hrule
\end{figure}

Manual corrections were made for the Armenian Catholic Bible (which
has many out-by-one misnumbered verses), and also for the Vulgate where
some verses are coalesced.

Each Bible scrape collected the language code for the translation,
which is a near match to ISO639-3, except that languages can have
variants based on geography or orthography.  For example, there are
separate translations for {\tt por} and {\tt por\_pt} (Portugese in
different countries); and
{\tt shu} and {\tt shu\_rom} (Chadian Arabic in traditional script or Roman
script). These are stored as separate languages within LEAFTOP.
Each ISO639-3 code (and variant) is connected to a language name and
geography using a Wikidata extract.

\subsection{Vocabulary Extraction}



The vocabulary extraction took 26,000 CPU hours,
which was parallel-processed
on a 64-cpu ARM processor in Amazon Web Services\footnote{A {\tt c6g.16xlarge} spot instance.}.

The vocabulary extraction algorithm is given in Algorithm
\ref{alg:vea}.  Note that it inefficiently calculates results
for all tokenisation methods --- unigrams (single words), unitokens (single unicode points) and quadgrams.
On completion, algorithm
\ref{alg:writtenstructure}\label{alg-written-structure-writeup} was run to identify which tokenisation
method is the most appropriate for the language, and other results are discarded.

In essence Algorithm \ref{alg:vea} runs a one-sided binomial test
asking whether a word or token in the target language appears
improbably often in the same verses as a Koine Greek lemma; the word
or token that is found most improbably often (having the lowest
$p$-value from the binomial test) is declared to be the best
translation. If there is a tie for least-likely-to-be-a-chance result,
no word is chosen.

\subsubsection{Example}

Consider how \textgreek{μνημεῖον} (tomb, grave)
is translated in the World English Bible.
 \textgreek{μνημεῖον}
 appears
 in 35 distinct verses in the gospels in various forms;
 it appears in the
nominative singular in only two verses: John 19:41 and John 19:42\footnote{{\it [41]
    Now in the place where he was crucified there was a garden. In the
    garden was a new tomb in which no man had ever yet been laid. [42]
    Then because of the Jews' Preparation Day (for the tomb was near
    at hand) they laid Jesus there.} --- World English Bible}.

The scraping process successfully captured 2862 verses,
but missed two of these 35
\textgreek{μνημεῖον} verses. Setting aside the other 31 verses
because they have the lemma \textgreek{μνημεῖον} in some
other case or number leaves 2831 verses from which we can calculate
a baseline probability-of-appearance for an English unigram.

Focussing on the two verses with the nominative singular
of \textgreek{μνημεῖον}, the World English Bible translation shows
52 unigrams (including punctuation) of which 38 are distinct: unigrams
such as ``tomb'', ``laid'', ``garden'' and ``the''.

The unigram ``tomb'' appears 6 times in the 2831 verses, giving it a
baseline probability of $2.2*10^{-3}$ of appearing in a random gospel
verse; ``garden'' appears 4 times, giving it a baseline probability of
$1.4*10^{-3}$; ``laid'' appears 29 times (baseline probability $1.0*10^{-2}$).
At the other extreme, the unigram ``the'' appears 1,916
times, for a baseline probability of 0.68. It is unsurprising when
``the'' appears in a verse, and very surprising when ``garden'' does.

The unigram ``tomb'' appears in both John 19:41 and John 19:42, as do the
unigrams ``the'' and ``laid'', but ``garden'' only appears in the
first of these verses.  We can then perform a one-sided binomial test
for each unigram. For ``tomb'', $B(2,2,2.2*10^{-3}) = 4.5*10^{-6}$;
``laid'' $B(2,2,1.4*10^{-3}) = 1.04*10^{-4}$; ``garden''
$B(1,2,1.4*10^{-3}) = 2.8 * 10^{-3}$; ``the''
$B(2,2,0.68) = 0.46$.  From this we can conclude that a valid
translation for \textgreek{μνημεῖον} into English is ``tomb''.

\subsubsection{Short-comings and failures}

An obvious problem with this approach is that languages that inflect nouns with
a case system that is substantially different to Koine Greek's ---
such as Arabic --- are handled quite poorly, since there will be many
distinct forms in the target language ``competing'' to be the best
translation for each Koine Greek lemma.

A more subtle problem arises
when more than one lemma translates into the same word in a target language
(such as ``fish'' in English being a translation for
\textgreek{ἰχθύς} and \textgreek{ὀψάριον}), since this alters the baseline
appearance probability. The binomial test is very sensitive to changes
in this baseline, and where there are mistakes in the LEAFTOP dataset, this
is often the root cause.

\subsubsection{Confidence score}

\begin{table}
\begin{tabular}{l | p{26mm} | r | r}
\textit{Lemma} & \textit{English} & \textit{correct\%} & \textit{Rank} \\
\hline
\textgreek{προφήτης} & prophet & 98.4 & 161 \\
\textgreek{ὕδωρ} & water & 95.9 & 160 \\
\textgreek{ἔτος} & year & 95.5 & 159 \\
\textgreek{ἄνεμος} & wind & 94.9 & 157 \\
\textgreek{πρόβατον} & sheep & 94.9 & 157 \\
\textgreek{παραβολή} & parable & 92.2 & 156 \\
\textgreek{χείρ} & hand & 91.4 & 155 \\
\textgreek{ἄρτος} & bread & 91.3 & 154 \\
\textgreek{θεός} & God & 91.3 & 153 \\
  \textgreek{ἱμάτιον} & coat & 90.3 & 152 \\
  \end{tabular}
  \caption{Top 10 lemmas most likely to be extracted correctly}
  \label{hardest}

  \medskip
  \hrule
\end{table}

These p-values from the binomial tests can be remarkably small -- the
median p-value across all languages for translating \textgreek{θεός}
(God) is $4.46*10^{-17}$%
, so it is more convenient to work in terms of the negative base-10 log of the
p-value.

The ratio of this negative log p-value of the best word to
the second best word is recorded in the LEAFTOP database as the
confidence score\label{confidence-score-definition}.
In the \textgreek{μνημεῖον} example, the next nearest alternative
to ``tomb'' is ``laid''; the ratio between their
log p-values is $1.3$.

As discussed in section \ref{improve-accuracy}, the confidence score
is a useful (but not sufficient) predictor of whether the vocabulary
is correct.

Figure \ref{mean-vs-standdev-confidence} shows the mean
and standard deviations of the confidence scores for each lemma.
Words like \textgreek{θεός} (God),
\textgreek{μαθητής} (disciple) and \textgreek{χείρ} (hand) are usually
easy to identify in most target languages, which is unsurprising as
they are very commonly used in the gospels and are unlikely to be
paraphrased. These have very high confidence scores.

Conversely, words like \textgreek{λίτρα} (a unit of measure),
\textgreek{σταφυλή} (grapes) and \textgreek{λύκος} (wolf) are usually
the hardest to identify, suggesting that translators were either
unable to be consistent in the way that they translate these terms or
that these terms regularly appear as part of a repeated multi-term
phrase.

\textgreek{ὀδούς} (tooth) is either easy to extract if the translators
translated Matthew 5:38\footnote{{\it You have heard that it was said,
    ``An eye for an eye, and a tooth for a tooth.''} --- World English
  Bible} very literally or nearly impossible to extract otherwise.
Similarly the confidence in extracting \textgreek{λύχνος} (lamp) is
heavily influenced by the translators' choices in Luke
12:35\footnote{{\it Let your waist be dressed and your lamps burning.}
  --- World English Bible}.

245 language codes have more than one translation available.  For
these languages, a consensus-by-vote for each lemma is taken based on
the results from the Bible versions in that language. The confidence
scores are multiplied\footnote{An additive model is also being
  investigated.} and stored as {\tt cumulative\_confidence} in the
LEAFTOP extracts for each language. Where there is a tie for the best
word, nothing is chosen.  For languages without a second translation,
a pseudo-consensus (the answer derived from the sole translation) is
used.

\section{Results}\label{results}

\begin{table}
\begin{tabular}{l | p{26mm} | r | r}
\textit{Lemma} & \textit{English} & \textit{correct\%} & \textit{Rank} \\
\hline
\textgreek{τροφή} & food & 37.5 & 8 \\
\textgreek{κλῆρος} & lots (casting of lots), inheritance & 37.5 & 8 \\
\textgreek{τράπεζα} & table & 37.5 & 8 \\
\textgreek{κοιλία} & womb, stomach, source of feelings and emotions & 36.4 & 7 \\
\textgreek{σταφυλή} & grapes & 35.3 & 6 \\
\textgreek{ἀρχή} & beginning & 32.4 & 5 \\
\textgreek{βρέφος} & babies & 31.1 & 4 \\
\textgreek{ὀφειλέτης} & debtor & 29.6 & 3 \\
\textgreek{πλήρωμα} & fullness & 25.0 & 2 \\
\textgreek{στάχυς} & head of grain & 21.9 & 1 \\
  \end{tabular}
  \caption{Bottom 10 lemmas least likely to be extracted correctly}
  \label{easiest}
  \medskip
  \hrule
\end{table}

The LEAFTOP dataset has 625,351 distinct words in 1502 different languages. 22
of those languages are variant forms of some other language (e.g.
{\tt zho\_tw}, {\tt urd\_dv}).
Taking each language and considering each Koine Greek lemma as a concept,
there are 239,156 distinct records.

\subsection{Evaluations of correctness}

The numbers in Section \ref{results} include words that
are incorrect. The error count is hard to obtain.  Randomly sampling
from 1,480 languages would have been impractical, since there would be
a high probability of landing on a language that has a very small
number of speakers, or is extinct.

Instead, the approach taken was to group by language geography, find
freelancers in the appropriate part of the world, and pay for them
either to check the extracted vocabulary themeselves, or to find
speakers of regional languages who could do this. This was only partly
successful; we were unable to find freelance translators for
any South American or North American indigenous language. Only one
Australian Aboriginal language has been checked, and that wasn't even
from one of the larger language families. The list of languages (and
the language family associated with them) is shown in Table
\ref{language-families-evaluated}.

The evaluations themselves have errors, which are preserved as-is in
the LEAFTOP evaluations data. Examples that we noticed include
French \textit{pains} being marked as incorrect for \textgreek{ἄρτος}
(bread), and Swahili \textit{Mungu} being marked as incorrect for
\textgreek{θεός} (God). There are likely to be more.

A chart summarising the evaluations done by the translators is shown
in Figure \ref{proportion-correct}. In total they checked 10,464
distinct words, corresponding to 5,120 (language, lemma) combinations;
respectively approximately 1.7\% and 2.1\% of the total vocabulary.

An interesting accident happened with Chadian Arabic. Due to a
miscommunication by the first author, the evaluator checked the translations
for both {\tt shu} (60.4\% correct) and {\tt shu\_rom} (59.2\%
correct) --- the latter being Chadian Arabic written in Romanized
letters. While it is premature to assume that LEAFTOP works equally
well across different writing systems, this result does hint at that.

\subsection{Approaches for improving vocabulary accuracy}\label{improve-accuracy}

\begin{figure}
  \includegraphics[scale=0.35]{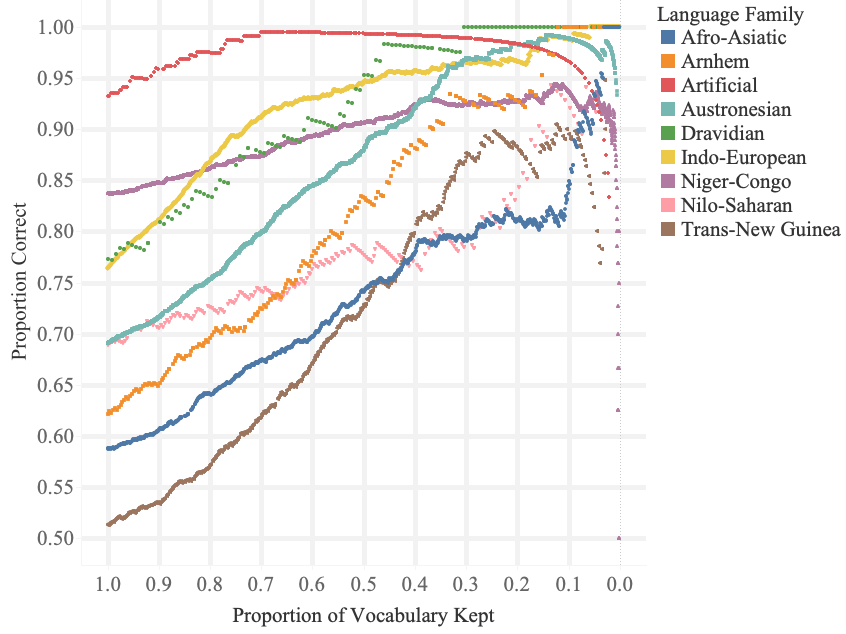}
  \caption{Trade-off curve of correctness vs size --- altering the
    confidence threshold cut-off generally increases the probability
    of having correct vocabulary}
  \label{roc-like}

  \medskip
  \hrule
\end{figure}

Linguists working with the LEAFTOP data may be willing to trade off
a smaller vocabulary for higher accuracy.

As shown by the trend line in Figure \ref{hardness}, confidence
doesn't fully explain accuracy. Even when the regression is against
the rank of the proportion correct, the $R^2$ is still only 0.83; so
even a non-linear monotonic relationship is not fully explanatory.
But in general, higher confidence scores are predictive of the higher
accuracy.

By setting a minimum confidence cut-off threshold, it is possible
to have a smaller data set that has higher accuracy.
Figure \ref{roc-like} shows the trade-offs that are possible. 

Considering Figure \ref{hardness} again, words like {\it wind}, {\it
  disease} and {\it finger} are disproportionately likely to be
extracted correctly, and conversely, religious terms such as {\it
  prophet}, {\it God} and {\it parable} are much harder to extract
correctly given how often these words appear in the Gospels --- it is
common for the LEAFTOP algorithm to mistake the word for God (for
example) in an usual case or number.

This offers an alternative approach, which is to filter out by
vocabulary.  Table \ref{hardest} lists the lemmas that are
the most likely to be correct, and Table \ref{easiest} lists
the lemmas that are most likely to be incorrect.

Finally, it is rare for the singular and plural of a word to differ
substantially, but there are many lemmas in the LEAFTOP database where
the extracted singulars and plurals differ.  \textgreek{χρόνος} (time)
is translated into German as {\tt Zeit} in the singular (which is
correct), and as {\tt l\"angere} (``longer'') in the plural. An
obvious filter that could be implemented for alphabetic languages is
to count the number of letter sequences in common and find a threshold
below which it is unlikely to be a correct translation; a Levenshtein
distance (or equivalent) could also be used. A more sophisticated
filter could be created by building a machine learning model to
predict the plural from the singular and to discard lemmas where there
is a mismatch. We are working on this latter approach.


\section{Exploratory Tools}

\begin{figure}[t]
  \begin{center}
    \framebox{\includegraphics[scale=0.25]{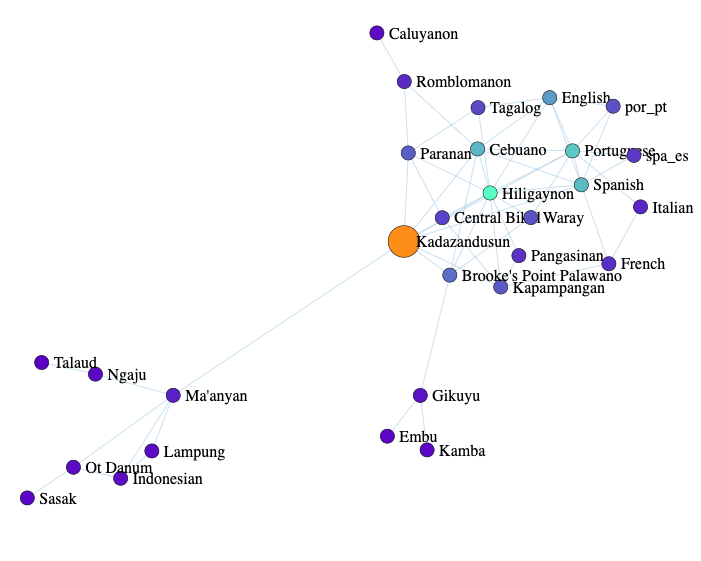}}
    
    \medskip
    
    \framebox{\includegraphics[scale=0.21]{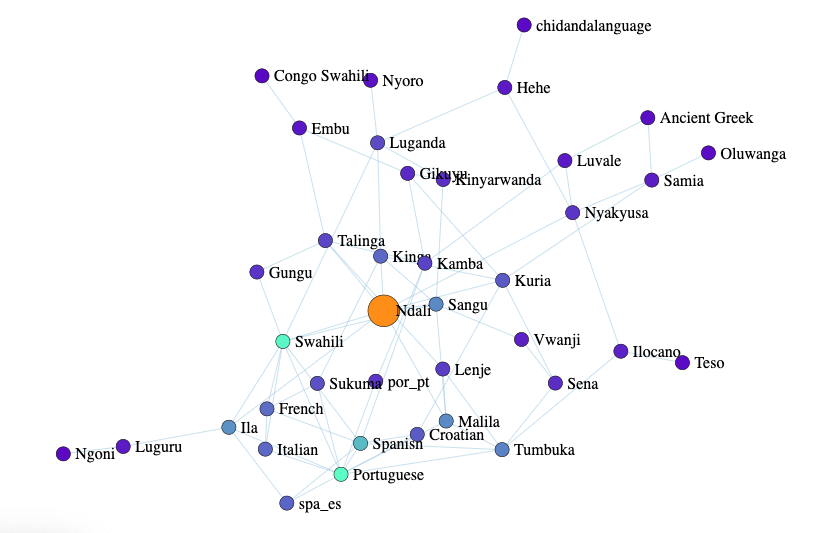}}
  \end{center}
    \caption{Screenshots of the LEAFTOP explorer}
    \label{explorer}
    \medskip
    \hrule
\end{figure}

The last component of the LEAFTOP database is the explorer. This is
an interactive set of web pages for showing the relationships between
different languages. The idea behind it is that if two languages
are related, then the challenges faced by translators should have been
similar --- words and concepts  that do not map
nicely from Koine Greek to one language should also be hard to
map into a related language. Likewise, concepts that have straightforward
mappings, should also be straightforward in a related language.

This is of course a vast simplification of a much more complex system
and the goal is to explore the limitations of this toy model.

This relatedness of two languages is quantified in the LEAFTOP dataset
by the Spearman correlation statistic between the confidence scores of
the two languages --- for each lemma the confidence score from
language 1 is the $x$ value and the confidence score for language 2 is
the $y$ axis. The approximately-160 data points can then be
correlated.

To visualise these correlation statistics and make an interesting
interactive demonstration of this data, the first author created a D3.js \cite{d3_for_lrec}
force simulation model.  Each language is modelled as a ball
connected to other languages by a spring. The strength of the spring
is proportional to the correlation. To simplify the user interface,
only the languages with the closest and strongest springs are shown to
the user.  Sample outputs are shown in Figure \ref{explorer}.

The results are simultaneously disappointing and exciting. Many
languages are connected to each other correctly.  Unfortunately, the
languages spoken by European missionaries and translators correlate
very strongly with many target languages in completely different
language families --- e.g.  the confidence scores of Spanish and
Hiligaynon are highly correlated across lemmas. Possible causes for
this could be an implicit bias by translators; or could be related to
the introduction of new vocabulary from the missionary's native
language substituting for vocabulary that didn't exist previously; or
it could simply be random noise.

\section{Conclusion}

The LEAFTOP dataset is an extremely large collection of nouns across
many different languages, with a measured accuracy across a variety of
language families. It has been used for building multilingual
pluralization models and for language exploration. There are
straightforward extensions that could be done to improve
its accuracy and coverage.

The source code for creating the dataset is
\url{https://github.com/solresol/thousand-language-morphology},
and
the final outputs (the dataset itself) are in
\url{https://github.com/solresol/leaftop}.

\section{Acknowledgements}

The authors would like to thank Daniel Everett and Mat\'ias
Guzm\'an Naranjo for their support, and acknowledge the contributions
of the freelance translators (some of whom have requested anonymity)
who have checked the translations: Benazir Bhagad, Eleanor M. Mendoza,
Maureen Y. Ong, Wewalage Roshan Chanaka Perera, Rim Sayed, Ferdaous
J., Eric Ojecty, Farah Taymour, Owembabazi Don, Auma Sharot, Okotoi
Ruby, Okullo Joel, Bwambale Hamza, Frencelin Laurice, Sidime Amadou,
Chimankpa Stanley, Moussa Keita, Paul Malanou, Uhtman Alake and
especially Bradley Gewa for his tireless work finding an Australian
Aboriginal language translator and all the translators for languages
in New Guinea.

\section{Bibliographical References}\label{reference}

\bibliographystyle{lrec2022-bib}
\bibliography{../bibliography.bib}


\end{document}